\documentclass[conference]{IEEEtran}
\IEEEoverridecommandlockouts
\usepackage{cite}
\usepackage{amsmath,amssymb,amsfonts}
\usepackage{algorithmic}
\usepackage{graphicx}
\usepackage{textcomp}
\usepackage{xcolor}
\usepackage{float}
\usepackage{gensymb}
\usepackage{siunitx}
\usepackage{hyphenat}
\def\BibTeX{{\rm B\kern-.05em{\sc i\kern-.025em b}\kern-.08em
    T\kern-.1667em\lower.7ex\hbox{E}\kern-.125emX}}

\usepackage[normalem]{ulem}
\usepackage{color, soul}
\sethlcolor{yellow}
    
\begin{document}
\title{Design and Fabrication of a Spin Coater with In-Situ Optical Measurement for Soft Thin Films \\
\thanks{The Japan Steel Works, LTD.}
}

\author{\IEEEauthorblockN{1\textsuperscript{st} Daniel Gliksberg}
\IEEEauthorblockA{\textit{Mechanical Engineering} \\
\textit{Massachusetts Institute of Technology}\\
Cambridge, MA, US \\
gliksber@mit.edu}
\and
\IEEEauthorblockN{2\textsuperscript{nd} Jiajie Qiu}
\IEEEauthorblockA{\textit{Mechanical Engineering} \\
\textit{Massachusetts Institute of Technology}\\
Cambridge, MA, US \\
qiujj@mit.edu}
\and
\IEEEauthorblockN{3\textsuperscript{rd} Jun Suzuki}
\IEEEauthorblockA{ \textit{The Japan Steel Works, LTD.}\\
Shinagawa, Tokyo, Japan \\
jun\_suzuki@jsw.co.jp}
\and
\IEEEauthorblockN{4\textsuperscript{th} Kamal Youcef-Toumi}
\IEEEauthorblockA{\textit{Mechanical Engineering} \\
\textit{Massachsetts Institute of Technology}\\
Cambridge, MA, US \\
youcef@mit.edu}
}

\maketitle

\begin{abstract}
Spin coating is widely used for fabrication of thin polymer and elastomer films, yet reliable thickness verification of highly compliant materials remains challenging due to deformation from contact-based measurements and the cost and complexity of conventional optical metrology. Accurate thickness control is especially critical in soft elastomer applications such as dielectric elastomer actuators (DEAs), where mechanical and functional performance scales strongly with film thickness. This work presents a low-cost, primarily 3D-printed benchtop spin coater with an integrated, minimally deforming optical thickness measurement system for soft-film fabrication workflows. The system is designed to manufacture films between 50 and 300 microns thick with repeatability within 10~microns. Thickness is measured in-situ by tracking displacement of a reflected laser beam via quadrant photodetector, avoiding significant deformation. Optical geometry, sensor linearity constraints, and structural validation via finite element analysis are discussed. Experimental validation using calibrated metal shims demonstrated a thickness resolution of 3.6-3.7~microns and best-case measurement repeatability of 13~microns (95 percent confidence interval). The platform repeatably produced silicone films within 9~microns of target thickness, demonstrating that accessible optical metrology can be integrated into a low-cost spin coating system for practical, thickness-controlled fabrication of compliant thin films without specialized industrial instrumentation.
\end{abstract}


\section{Introduction}

Spin coating is a widely used method for forming thin polymeric films, where centrifugal force drives radial outflow and progressive film thinning on a rotating substrate. Early analyses modeled the thinning of a viscous liquid layer on a rotating disk and showed how rotation promotes uniformity under idealized Newtonian flow assumptions~\cite{emslie_flow_1958}. Several following studies extended the framework to practical polymer systems, examining the effects of different parameters on final film thickness and uniformity~\cite{meyerhofer_characteristics_1978, flack_mathematical_1984, lawrence_mechanics_1988}.
While these models provide a useful framework for selecting spin parameters and predicting thickness, achieving reliable thickness in soft-film fabrication still requires direct thickness measurement.

The ability to accurately measure film thickness is particularly important in soft elastomeric systems where mechanical and functional performance scales strongly with thickness. For example, in dielectric elastomer actuators (DEAs), actuation strain is governed by the electric field applied across the elastomer, which is inversely proportional to film thickness. As a result, small variations in thickness can lead to substantial changes in generated stress, deformation, and dielectric breakdown thresholds~\cite{muffoletto_anticipating_2013}. More generally, in soft polymer films, bending stiffness, tensile compliance, and load-bearing capacity all scale with thickness. Consequently, reliable thickness verification is a prerequisite for consistent mechanical performance and reproducible device behavior in soft-film manufacturing workflows.

Conventional thickness measurement tools, such as calipers and micrometers, rely on direct mechanical contact and are therefore poorly suited for thin, highly compliant films. Even modest measurement forces can induce significant elastic deformation in soft elastomers, leading to thickness errors that are comparable to or larger than typical fabrication tolerances.

While non-contacting optical thickness measurement techniques such as reflectometry, ellipsometry, and interferometric microscopy can achieve high resolution under controlled conditions, they often rely on well-defined optical properties, reflective interfaces, rigid substrates, or complex calibration procedures~\cite{heavens_optical_1960, fujiwara_spectroscopic_2007, de_groot_principles_2015}. In addition, many established optical metrology systems are designed for cleanroom or industrial environments and typically require removing the sample from the fabrication tool for measurement, limiting their practicality for benchtop soft-film manufacturing workflows. As a result, there remains a gap between the need for reliable thickness verification of compliant films and the accessibility of existing measurement approaches.

Commercial benchtop spin coaters typically achieve maximum rotational speeds of 6,000–8,000~RPM, enabling fabrication across a wide range of film thicknesses~\cite{preasion_spin_nodate, ossila_spin_nodate, chemat_chemat_nodate}. However, compliant film applications such as DEAs generally do not require such high speeds, allowing system design requirements to be relaxed and reducing dependence on high-precision, high-cost motors and motor drivers.

Commercial benchtop spin coaters, at time of writing, range in price from approximately \$1,000 for basic laboratory units to \$5,000–\$7,000 for programmable research-grade systems~\cite{preasion_spin_nodate, ossila_spin_nodate, chemat_chemat_nodate}. Spin coaters are rarely integrated with thin-film metrology tools, and instead require standalone measurement units, with high-end devices requiring manufacturer quotes for pricing~\cite{bruker_filmtek_nodate}. This cost barrier motivates the development of lower-cost alternatives capable of accurate manufacturing and measurement, at the expense of requiring the user to manufacture 3D printed parts and assemble the system themselves. Such devices, without integrated measurement systems, have been published before, but are susceptible to closed-source firmware that has since been pulled off the market~\cite{carbonell_rubio_maasi_2022, alex_important_2024}.

The primary contributions of this work are: (1)~the integration of fabrication and minimally-deflecting thickness measurement into a single, accessible benchtop system; (2)~{a single-point, material-independent} optical measurement approach suitable for highly compliant films where contact-based tools fail; (3)~quantitative validation of measurement accuracy and repeatability using calibrated reference samples; and (4)~demonstration of the system’s applicability to soft-film manufacturing without reliance on specialized industrial metrology equipment. Together, these contributions address a practical gap between soft-film fabrication and reliable thickness verification in research-scale manufacturing settings. 

\section{System Architecture}

The proposed system integrates fabrication and thickness metrology within a single benchtop system composed of three interacting subsystems: (1) a low-cost, 3D-printed spin coater; (2) a 3D-printed optical measurement system; (3)~a control and data acquisition framework. The combined system enables manufacturing and in-situ thickness measurement of compliant films, thereby avoiding handling-induced and contact-induced deformation caused by sample removal from the sample plate. The optical measurement subsystem infers thickness from the displacement of a reflected laser beam using a simple geometric-optical configuration, allowing minimally deforming measurement with micron~scale resolution. Because compliant films deform significantly under forces applied by common benchtop tools, such as calipers and micrometers, the system is designed to minimize measurement-induced loading and associated thickness deflection.

During operation, liquid precursor material is deposited and spun under closed-loop control. After curing at room temperature, the optical subsystem is used to measure thickness via beam displacement sensing, while the sample remains on the rotating plate. The voltage measurements from the optical sensor are then processed into a measured thickness.

Prior to fabrication, the measurement system is zeroed so that the beam spot lands in the center of the sensor's linear domain without a sample present. After curing, the lightweight mirror is reintroduced and beam displacement relative to this reference position is converted into a thickness measurement. This procedure also enables sequential measurement of multilayer films by re-zeroing after manufacture of each layer, which is particularly useful for DEAs.

This system is designed to manufacture films of thicknesses between 50 and 300~µm repeatably to within 10~µm. This implies a desired resolution below 5~µm, such that system resolution is at least 10 times finer than the smallest expected measurement. These requirements guided sensor selection, optical geometry, and structural design.

\section{Design}

\subsection{Mechanical Design}

\begin{figure}[bp]
    \centering
    \includegraphics[width=0.9\linewidth]{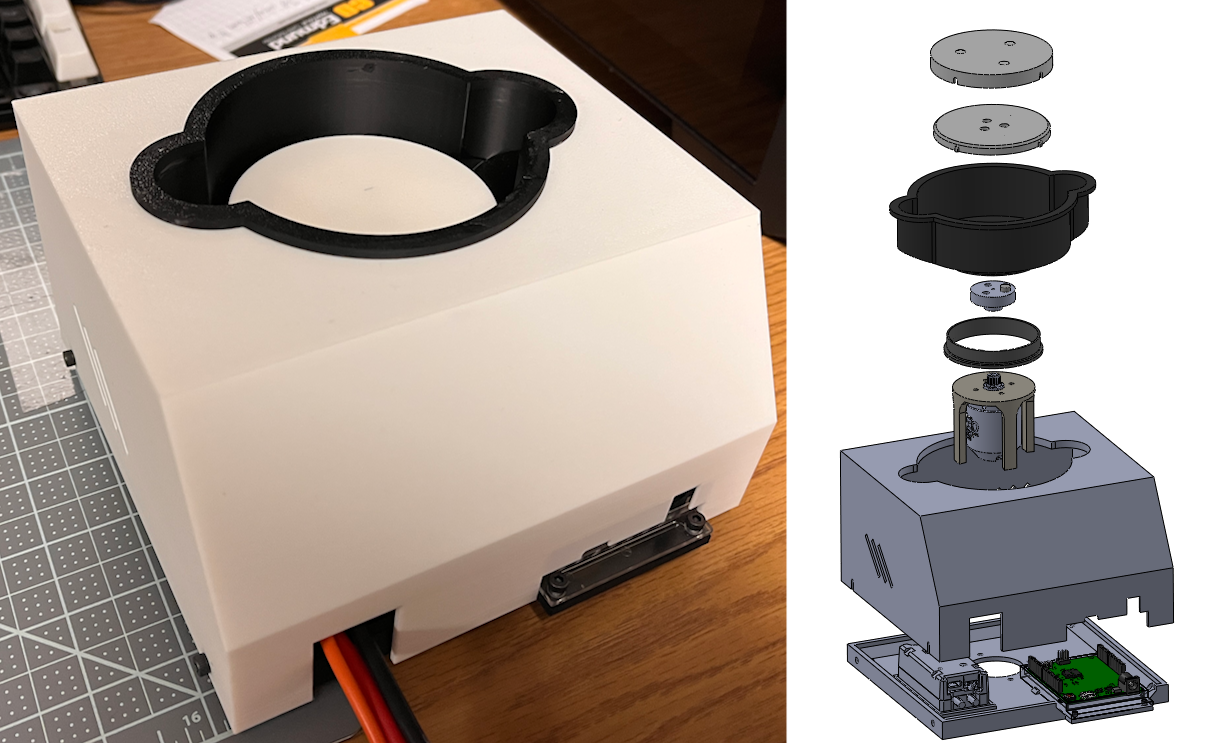}
    \caption{The assembled spin coater, with exploded view. Note that the SPI CAN Bus transceiver is not represented in the CAD model of the spin coater.}
    \label{fig:spinCoater}
\end{figure}

The majority of components for the spin coater proposed in this paper are 3D printed in PLA, with remaining components costing less than $\$200$ at time of publication. A full parts list for non-3D printed components is in Appendix~\ref{appendix:coaterParts}, and print settings for all 3D printed components are in Appendix~\ref{appendix:printSettings}. {The spin coater} is shown in Fig.~\ref{fig:spinCoater}.

\begin{figure}[tbp]
    \centering
    \includegraphics[width = .8\linewidth,height=3in,keepaspectratio]{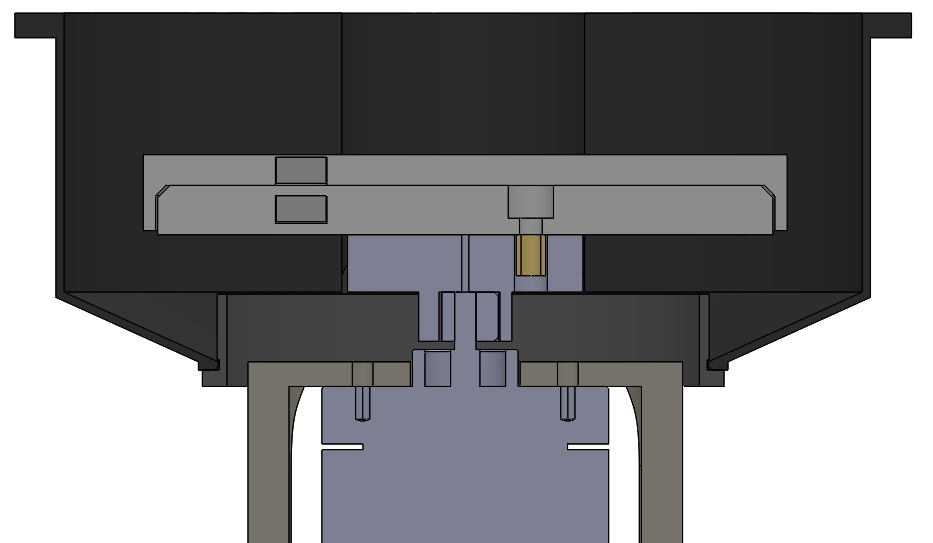}
    \caption{A cutaway view demonstrating the housing 'cup' geometry as well as the stackup of adapters between the motor and sample plate. }
    \label{fig:spinCoaterCutaway}
\end{figure}

The external housing of the spin coater is comprised of an outer shell, cup, and base, to which the motor controller, motor mount, and Arduino UNO are connected. The shell and cup protect the electronics from the spin coating material. The walls of the cup extend below and under the {3D printed} sample plate, and terminate at a central hole. This hole is required to fit the motor and sample plate, and thus a snap-in lip is added around the edge of this hole. This lip traps material that falls off the sample plate during use. So long as an excessive amount of material is not used, and the system is periodically cleaned, this lip will catch all excess material. A cutaway view of the housing cup is shown in Fig.~\ref{fig:spinCoaterCutaway}.

The motor has a pinion pressed onto its output shaft, to which a 3D printed adapter piece is pressed. The press-fit bore of this adapter is shaped to match the pinion, which is done to prevent slip as the PLA experiences stress relaxation because of pressure from the press fit. This motor adapter is then attached to the sample plate base via threaded inserts and countersunk screws. The sample plate base contains encased magnets to retain the sample plate and radial pins to prevent incorrect plate placement. This component is separate from the motor adapter because it must be installed after the housing cup is in place, and also because it presents a convenient location for shims to align the sample plate to the axis of rotation. The stackup of components between the motor and sample plate is shown in Fig.~\ref{fig:spinCoaterCutaway}. 

The sample plate itself is held in place magnetically via three neodymium magnets, matching the sample plate base, and aligned with four notches distributed around its outer edge. All sample plates were printed so that the sample contact area was against the print bed, which results in a more even surface than when printed as a free surface. This is only the case on a smooth print plate.

\subsection{Optical Measurement System}

\begin{figure}[htbp]
    \centering
    \includegraphics[width=0.75\linewidth]{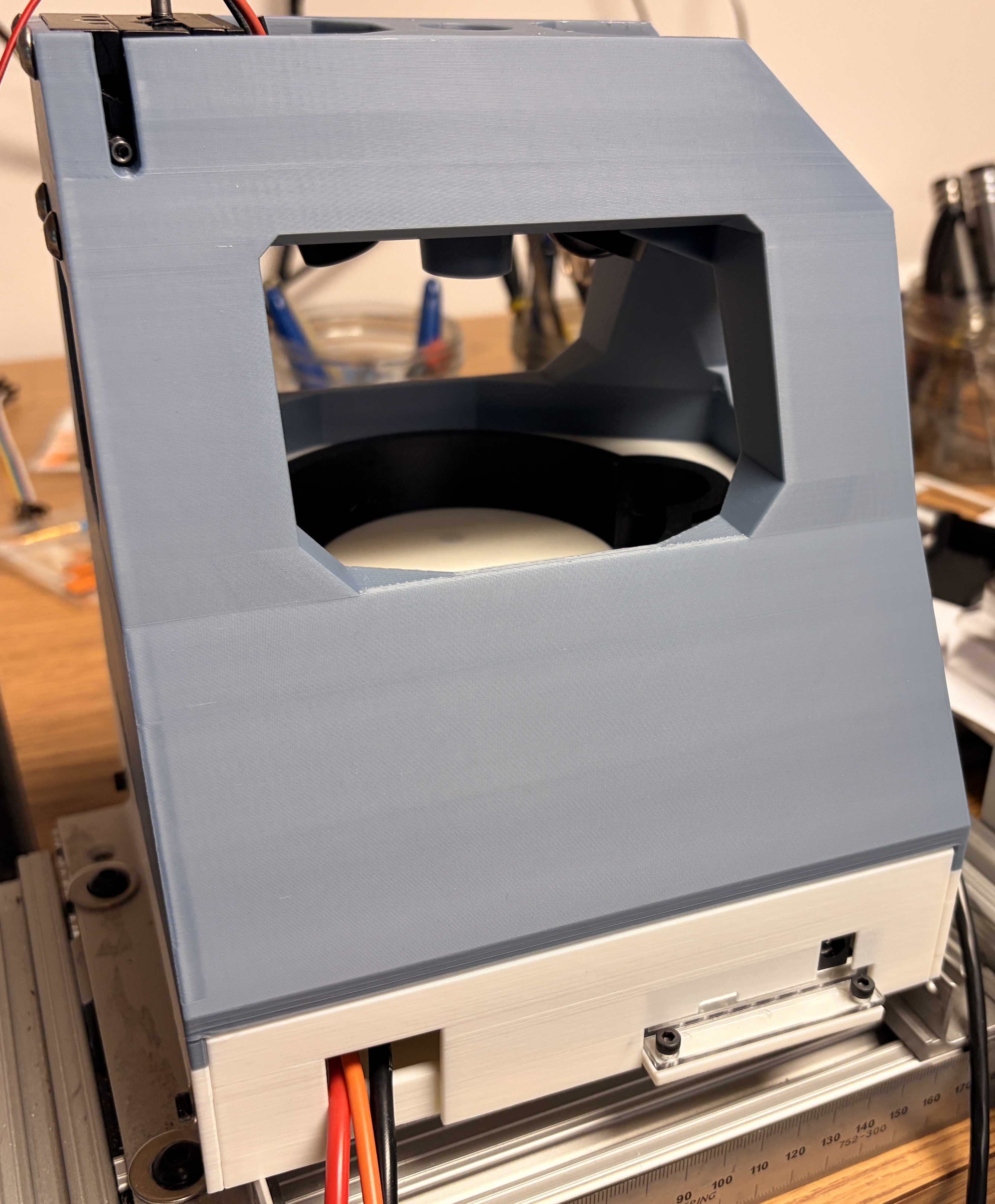}
    \caption{The measurement system installed on the spin coater. The 3D printed structure holding the optical measurement components, printed in blue, is referred to as the measurement frame.}
    \label{fig:measurementSystem}
\end{figure}

The chosen operating method of the spin coater's integrated measurement system is, at a high level, to track the position of a laser on an optical sensor before and after the presence of a sample. This position is then converted into a measured sample thickness. The measurement must be done without significantly deforming soft film samples, which are commonly optically clear or otherwise not reflective. A mirror must be placed {on top of} the sample {so that the beam may be reflected independently of sample material properties}, and this mirror must be sufficiently light so that the sample deflection under the added weight can be neglected in analysis. Fig.~\ref{fig:measurementSystem} demonstrates the assembled measurement system. 

Easily accessible and non-destructive benchtop measurement methods, such as calipers and micrometers, require applied forces that would significantly deform samples made of these materials. One such material, referred to here as Ecoflex10T, has a Young's Modulus of $14$~\unit{kPa} for strains less than 300\% when cured~\cite{jamali_development_2022-1}. The mirror was designed so that placing it in direct contact with material of similar Young's Modulus would cause minimal deflection. The deflection of the sample under the weight of the mirror can be estimated by treating the sample as a simple two-force member under load via~(\ref{eqn:axialLoading}). For this equation, $\delta$ is the total deflection of the sample, $F$ is the applied load, $E$ is the material Young's Modulus, and $A$ is the sample cross-sectional area.
\begin{equation} \label{eqn:axialLoading}
    \delta = \frac{FL}{EA}
\end{equation}
The surface area of the sample plate is $A~=~\num{4.83e-3}$~\unit{m^2}, the weight of the mirror is~$F~=~0.292$~\unit{N}, and an example~$L~=~0.3$~mm thick Ecoflex10T sample is assumed.~For such a sample, the weight of the mirror results in a total deflection of $\delta~=~1.29$~\unit{\micro\meter}. In contrast, the micrometer used to validate system measurements has a measurement force of 4.41~N. As per~(\ref{eqn:axialLoading}), for an evenly applied axial force of 4.41~N the same Ecoflex10T sample would deform by $\delta = 19.5$~\unit{\micro\meter}.

Many of the considerations made in the design process of the optical measurement system were driven by the chosen optical sensor, the Thorlabs PDQ80A Quadrant Sensor. This sensor imposes notable limitations on beam spot size, wavelength, and angle of reflection with the sample.

As per manufacturer specifications, the PDQ80A provides the best results at low optical powers, based on the sensor's responsivity at any given wavelength in its rated range. The peak responsivity of the sensor is at wavelengths around 950~\unit{nm}. Given this ideal wavelength, and the minimum and maximum optical power specifications for the sensor, the Thorlabs L895VH1 laser diode was chosen, operated at 3.6~\unit{V} in series with a $1.4~\text{k}\Omega$ current-limiting resistor. The laser diode was mounted on a fine-adjust stage to allow for system zeroing. From there, lenses were chosen to collimate the light emitted by the laser diode and to focus the reflected beam such that the beam spot diameter {at} the PDQ80A is 0.7~\unit{mm}.

\begin{figure}[tbp]
    \centering
    \includegraphics[width = .9\linewidth,height=3in,keepaspectratio]{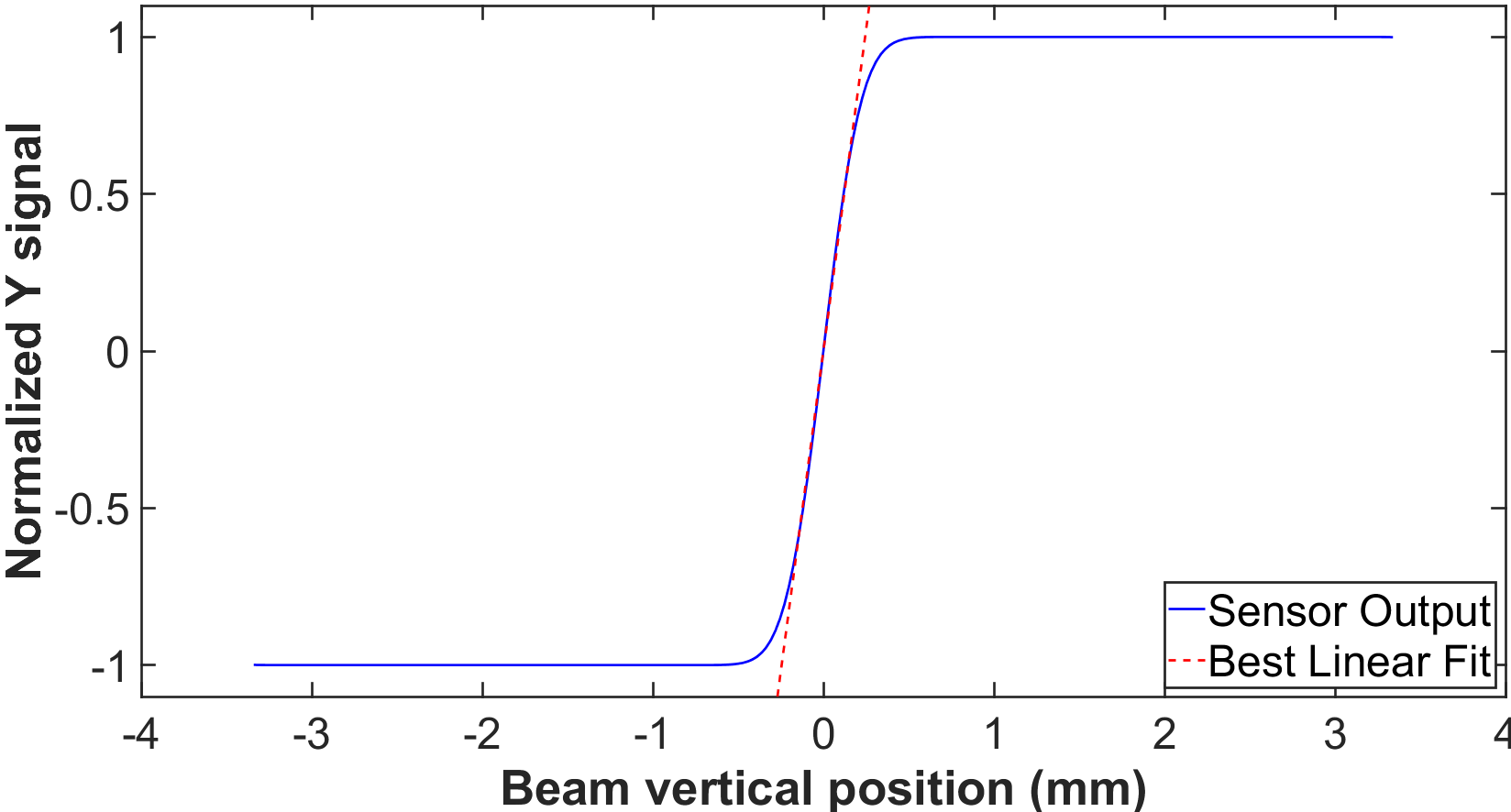}
    \caption{Simulated and normalized sensor Y signal response to a 0.7~mm diameter beam swept vertically across it, demonstrating the nonlinear response of the sensor when the beam center is too far from the sensor origin.} 
    \label{fig:Sensor}
\end{figure}

The response of the Thorlabs PDQ80A to a vertically swept beam spot of this diameter is shown in Fig.~\ref{fig:Sensor}. The observed nonlinearity drove the geometry of the measurement frame, because the desired measurement range of the system could not result in beam deflection beyond the linear domain of the sensor.

The angle of reflection between the laser and the mirror determines the range of sample thicknesses that fall in the linear domain of the sensor. The relationship between sample thickness $\tau$ and the beam motion detected by the sensor $y$ is related to the angle of reflection $\theta$ as follows:
\begin{equation} \label{eqn:beamDeflection}
    y = \frac{\sin(\pi - 2\theta)}{\sin(\frac{\pi}{2} - \theta)}\tau
\end{equation}
A sketch of the beam geometry is shown in Fig.~\ref{fig:beamGeometry}, and the derivation of this equation is in Appendix~\ref{appendix:beamProof}. 

\begin{figure}[bp]
    \centering
    \includegraphics[width = .9\linewidth,height=3in,keepaspectratio]{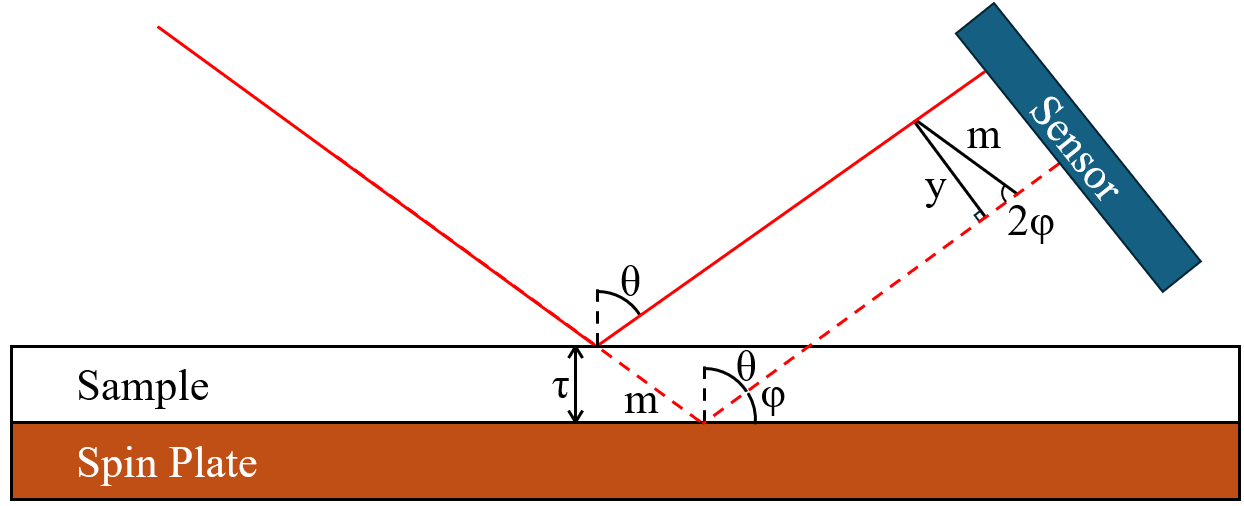}
    \caption{Beam paths before the presence of a sample (dashed) and after (solid). $\theta$ is the beam angle of reflection, $\phi$ is its complementary angle, $\tau$ is the thickness of the sample, $m$ is the extra beam travel length in the absence of the sample, and $y$ is the beam travel detected by the sensor. Note that this figure is simplified and omits the mirror used to reflect the beam for clarity.}
    \label{fig:beamGeometry}
\end{figure}

The beam angle of reflection is set by the measurement frame, and it was determined that, to maximize measurable sample thicknesses, the angle of reflection should be chosen so that the beam deflection is less than the thickness of the sample. For this reason, an angle of reflection of $\theta = 20\degree$ was chosen, resulting in a scaling factor of $0.684$ between sample thickness and beam spot deflection. Note that the sensor outputs voltage signals, and that the calibration process does not use this scaling factor explicitly

\subsection{Opto-Mechanical Integration}

The shape of the mirror used by the optical measurement system was chosen to match the internal profile of the spin coater cup. This aids in mirror alignment by ensuring that the mirror comes to rest centered on the sample, reducing the influence of the system operator on measurements.

Design validation of the measurement frame was performed using Finite Element Analysis (FEA) to quantify gravitational deflection, which would alter the beam path and introduce measurement error. For simulation, the frame geometry above the spin coater was simplified into a shell model with a maximum wall thickness of 1 mm. Only small assembly alignment indents were removed; these features were assumed to have negligible structural influence. The resulting shell contains no infill and vertical walls 1.2 mm thinner than the fabricated part, producing a conservative simulation of gravity-induced deflection. The assumed material properties for FDM printed PLA are described in Table~\ref{tab:PLA}. PLA was modeled as an orthotropic material. Note that, because of the variability of the FDM printing process, these numbers are rough estimates collated from multiple sources. {External effects such as ambient temperature and induced vibration are assumed negligible for the conditions required for DEA fabrication, but may be considered for high-performance materials.}

\begin{table}[htbp]
    \caption{Material properties of PLA used for simulation of measurement frame deflection under gravity.}%
    \begin{center}
    \begin{tabular}{|c|c|}
    \hline
    \textbf{Quantity}&\textbf{Value} \\
    % \cline{2-4} 
    \hline
    Density ($\textrm{g}/\textrm{cm}^3$)~\cite{noauthor_pla_nodate} & 1.24 \\
    Young's Modulus (XY) (MPa)~\cite{noauthor_pla_nodate}  & 2580 \\
    Young's Modulus (Z) (MPa)~\cite{noauthor_pla_nodate} & 2060 \\
    Poisson's Ratio~\cite{ferreira_experimental_2017} & 0.33 \\
    Shear Modulus (MPa)~\cite{ferreira_experimental_2017} & 1090 \\
    \hline
    \end{tabular}
    \label{tab:PLA}
    \end{center}
\end{table}

\begin{figure}[tbp]
    \centering
    \includegraphics[width = .9\linewidth,height=3in,keepaspectratio]{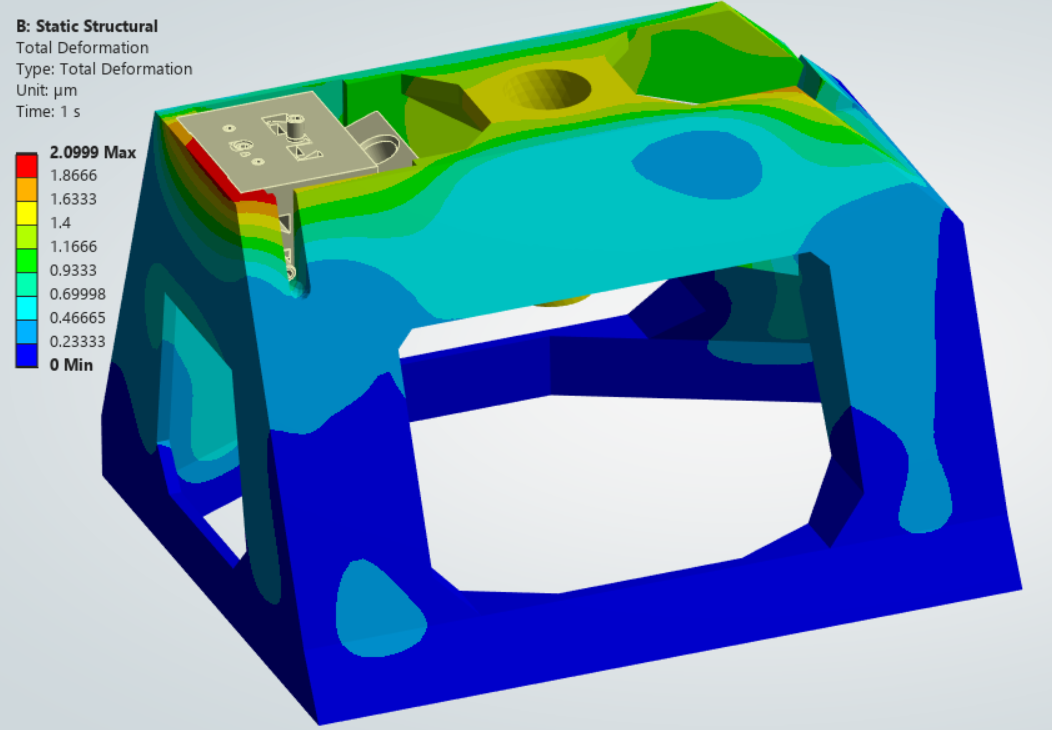}
    \caption{Simulation results for the deflection of the measurement frame under gravity, under conservative structural assumptions. The underside of the frame is fixed to ground. The peak total deflection is 2.10~\unit{\micro\meter}, at the center of mass of the combined linear stage and laser module.}
    \label{fig:gravityFEA}
\end{figure}

For the simulation, the fine-adjust stage was merged with the laser module. Both this module and the optical sensor were modeled as rigid bodies with uniform density. Their masses were set to 136.28~g and 71.95~g, respectively. The sensor connection to the frame was modeled as a bonded contact at all coincident surfaces. The merged laser object was only connected to the wall it is bolted to. Simulation results are shown in Fig.~\ref{fig:gravityFEA}.

Prior to experimental validation, the sample plate was aligned to the spin coater rotation axis by inserting shim stock between the motor adapter and sample plate adapter. Alignment was verified using a dial indicator measuring maximum vertical displacement at the sample plate edge over one revolution. The minimum achieved total deflection was 0.10 mm, corresponding to a 0.15\degree~misalignment between the sample-plate normal and rotation axis.

Although this angular misalignment affects spin coater operation, its influence during measurement validation was constant {because measurements were only taken when the sample plate was in the same angular position as in calibration}. {Any potential misalignment was therefore captured in the calibration conversion factor.}

\subsection{Control and Data Acquisition}

\begin{figure}[bp]
    \centering
    \includegraphics[width = .9\linewidth,height=3in,keepaspectratio]{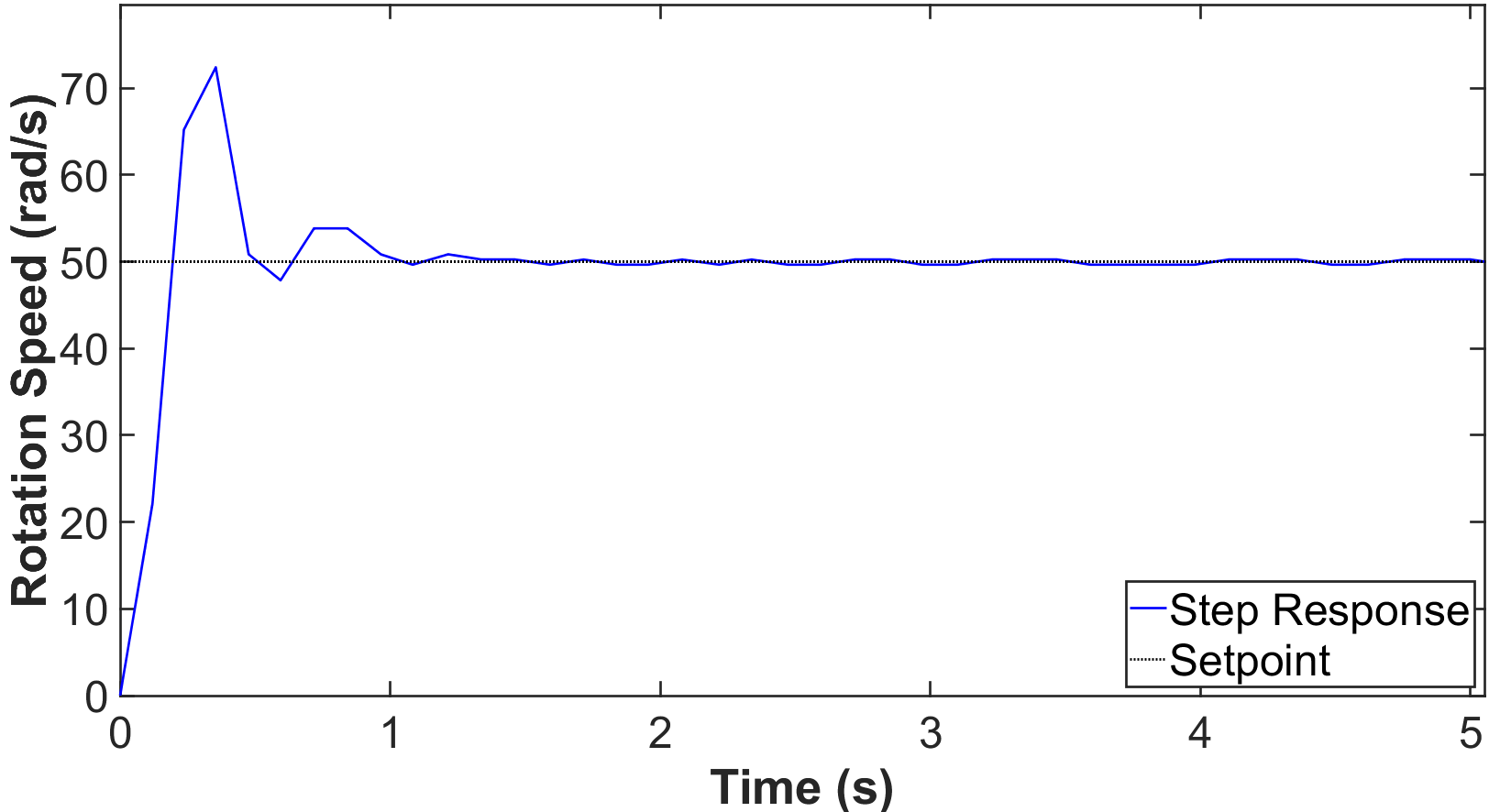}
    \caption{Step response of spin coater to a 50~rad/s setpoint. The rise time of the controller is 0.12~s, the 2\% settling time is 0.96~s, and the percent overshoot is 45\%. The sampling rate of REV Hardware Client, the program used to read the encoder data from the motor controller, is 8.5~Hz. The chosen motor, the REV NEO 550, has an encoder resolution of 42 counts per revolution, resulting in a velocity resolution of $1.27$~\unit{\radian/\second}. This manifests as the observed steady-state variation in measured velocity.}
    \label{fig:coaterStepResponse}
\end{figure}

The motor is operated via Arduino UNO with a SPARK MAX motor controller, and given 12~V at 2~A through the SPARK~MAX via benchtop power supply. The SPARK MAX can only run closed-loop control on external setpoints when communicating via CAN Bus, and thus a SPI CAN Bus module is required. Spin speed and duration are controlled via the Arduino UNO, which sends setpoints to the built-in velocity PID loop of the SPARK~MAX over the CAN connection. The velocity step response of the spin coater is shown in Fig.~\ref{fig:coaterStepResponse}, demonstrating the tuning of the PID controller and motor encoder resolution. The Arduino is operated by uploading the code, which automatically begins the spin coating process. The Arduino is powered via USB connection to a computer.

The duration and velocity setpoints are currently set by the user in code. In future iterations of this spin coater, a dedicated user interface to initiate spin coating and change these setpoints would be preferable.

The optical sensor is a quadrant detector that is read by a myRIO-1900 via its built-in 12-bit analog-digital converter (ADC) with a ±10~\unit{V} range. Three signals are read from the sensor: X position, Y position, and sum of all quadrants together. The position signals are defined as the difference in voltage outputs between horizontal and vertical semicircles of the sensor, respectively, and the sum signal is the sum of voltage signals coming from all four quadrants. The Y position signal is used to calculate sample thickness, the sum signal is used to normalize the signal so that measured thickness is not dependent on the overall optical power received, and the X position signal is only used for zeroing the system.

\subsection{Measurement Modeling and Calibration}

The system must be zeroed before sample manufacturing and measurement because the optical sensor response is only linear in a narrow range around the sensor origin, where the X and Y signals are 0~V for a non-zero optical power input (i.e. non-zero sum signal). This is done by placing the system mirror in contact with the surface to be taken as a reference and using the fine-adjust stage to move the laser until the X and Y position signals are zero.

To determine the conversion constant between Y signal voltage and sample thickness, calibration is required. This was done by first zeroing the system to the sample plate, {held at a constant reference position}. {A reference shim} was then inserted under the mirror, and the Y signal voltage was recorded as a 10~second time series. Using the average of the recorded voltage, and 5 averaged micrometer measurements of the reference shim, a conversion constant between voltage output and sample thickness can be determined. This process also determines the resolution of the system, because this conversion constant is applied to the voltage resolution of the myRIO's ADC. The voltage signal is averaged to prevent {signal noise} from influencing the calibration process.

The conversion constants determined for the metal shim experiment set the thickness resolution of the measurement system. The conversion constants used for the 0.2~\unit{mm} and 0.3~\unit{mm} calibration cases in Table~\ref{tab:Results} were 1.32~\unit{V/mm} and 1.35~\unit{V/mm}, respectively. Note that these constants are similar because both sample thicknesses fall in the linear range of the sensor. As a result of the measurement frame geometry, and the PDQ80A's voltage resolution of $4.88$~\unit{mV} for a 12-bit ADC with a 20~\unit{V} range (±10~\unit{V}), the thickness resolution for these two calibrations are 3.7~\unit{\micro\meter} and 3.6~\unit{\micro\meter}, respectively. Note that, as per Fig.~\ref{fig:gravityFEA} and~(\ref{eqn:axialLoading}), the combined deflection of the measurement frame under gravity and the sample under the weight of the mirror is lower than the resolution of the sensor in both cases. Therefore, it is reasonable to neglect these two effects in analysis. The system resolution is less than the design goal of 5~\unit{\micro\meter}, satisfying it.

\section{Performance Characterization}

\subsection{Experimental Verification of Measurement}

In order to determine the accuracy and repeatability of measurements taken with the proposed measurement system, measurements of ring-shaped stainless steel shims of three nominal thicknesses were taken and compared to micrometer measurements. Because there is variation in the thicknesses of each individual shim, these measurements can be interpreted as representative of system accuracy in a narrow range around each nominal thickness. 

All experiments were done in a dark room to minimize the effect of ambient light on sensor readings.

The measurement system was calibrated twice to sample shims of different nominal thicknesses: one that was nominally 0.2~mm thick, and one that was nominally 0.3~mm thick. After each calibration, measurements were taken of four shim samples from each nominal thickness (0.2~mm, 0.3~mm, and 0.5~mm). Each sample was measured five times, both by the measurement system and by micrometer. Each measurement from the optical system was the average over 10 seconds of time-series sensor data, in the same manner as the measurements for calibration. In between system measurements, the mirror was re-seated on the sample by lifting it 1~\unit{cm} vertically and releasing it, in order to simulate the variability of mirror placement in a real use case. Each micrometer measurement was taken at a different point on the shim. The system was zeroed between each sample, but not between each measurement.

In order to determine measurement accuracy and repeatability, the following steps were taken to process measurement data: (1)~the time-series data for each system measurement was averaged; (2)~all five optical system measurements for each sample were averaged, all while propagating uncertainty; (3)~the five micrometer measurements for each sample were averaged; (4)~the measurement error for each sample was calculated as the difference between the system and micrometer measurements; (5)~these errors were averaged within each nominal thickness category to give a final system error per category. This process was repeated for both calibrations, with results shown in Table~\ref{tab:Results}. The resolution of the system is, as discussed previously, determined by the calibration process.

\subsection{Silicone Fabrication Demonstration}

In order to verify the functionality of the combined spin coater and measurement system, 0.3~\unit{mm} thick samples of Ecoflex10T were manufactured and measured. Ecoflex10T has a density of $\rho = 987 \; \unit{kg.m^{-3}}$ and viscosity of $\eta = 6.7$~\unit{Pa.s}~\cite{jamali_development_2022-1}. The formula used to determine spin coater operation parameters is referenced from Emslie, A. G., Bonner, F. T., and Peck, L. G~\cite{emslie_flow_1958}. For this formula, $h$ is the final sample thickness, $h_0$ is the initial thickness of deposited material, $\omega$ is the rotational speed of the spin coater, and $T$ is the duration of rotation~\cite{emslie_flow_1958}. There are two parameters that control the operation of the spin coater: spin speed $\omega$ and spin duration $T$. Therefore, $\omega$ was set so that $T$ would be at least an order of magnitude larger than the spin coater settling time determined from Fig.~\ref{fig:coaterStepResponse}, 0.96~s.

\begin{equation}\label{eqn:spinThickness}
h = \frac{h_0}{\sqrt{1 + \frac{4 \rho \omega^2h_0^2T}{3 \eta}}}
\end{equation}

For $\omega = 50$~\unit{rad/s} and $h_0 = 3.3$~\unit{mm}, a spin duration~$T$ of $22.45$~\unit{s} is required to create a sample 0.3~\unit{mm} thick.

For manufacturing, 5~\unit{g} of uncured material was deposited in the center of the sample plate. This material was allowed to settle for 10~\unit{s} before rotation began. Three samples were manufactured in this way and allowed to cure at room temperature overnight. Each was measured 5 times, with the mirror re-settled between each measurement consistent with the method used for the metal shims. Because the resultant material has such a low Young's Modulus ($14$~\unit{kPa}), however, a direct micrometer measurement cannot be done. The material will deform significantly before the micrometer clutch is tripped. Measurements were instead taken by first zeroing the micrometer to a stack of two empty, face-to-face sample plates. Sample thickness was then measured by placing an empty sample plate face down on the sample, not yet removed from its sample plate, and measuring the resultant stackup. This is still less than ideal, because the silicone will still elastically deform as a result of the measurement force of the micrometer, but with the load more spread out the measurement will be a closer match to the axial loading estimate derived from~(\ref{eqn:axialLoading}). The results of this test are shown in Table~\ref{tab:SiliconeResults}.

\section{Results \& Discussion}

\subsection{Experimental Verification of Measurement}

Experimental validation using calibrated metal shims demonstrated that, within the linear operating range of the optical sensor, the system achieves a thickness resolution of approximately 3.6-3.7~\unit{\micro\meter}, with measurement errors generally smaller than associated uncertainty bounds and best-case measurement repeatability of ±13~\unit{\micro\meter} (95\% confidence interval) when calibrated near the intended measurement thickness. These results validate the underlying geometric-optical measurement approach and establish a practical operating regime for accurate measurements. Performance degradation observed for thicker samples is consistent with known sensor linearity limits and suggests improvement through refinement of optical geometry, adjustment of beam angles, or expanded sensing configurations.

As shown in Table \ref{tab:Results}, average error was lowest when the calibration shim matched the nominal thickness of the measured samples. For most measurements 95\% confidence intervals exceeded average error, indicating no discernible measurement bias within the sensor’s linear domain. Confidence intervals ranged from 13–29~\unit{\micro\meter} near the calibration thickness and were notably larger for other nominal thicknesses. These results indicate that calibration should be performed with a reference sample comparable in thickness to {the intended samples}.

\begin{table*}[htbp]
    \caption{Average errors for each nominal thickness for each calibration shim sample, with 95\% confidence intervals, defined such that negative errors indicate system measurements lower than micrometer measurements.}%
    \begin{center}
    \begin{tabular}{|c|c|c|c|}
    \hline
    \textbf{Calibration Sample} & \textbf{0.2 mm error (mm)} & \textbf{0.3 mm error (mm)} & \textbf{0.5 mm error (mm)} \\
    % \cline{2-4} 
    \hline
    0.2 mm & 0.007 ± 0.029 & 0.021 ± 0.056 & 0.013 ± 0.083 \\
    0.3 mm & 0.026 ± 0.069 & -0.004 ± 0.013 & -0.11 ± 0.10 \\ 
    \hline
    \end{tabular}
    \label{tab:Results}
    \end{center}
\end{table*}

The ±13~\unit{\micro\meter} confidence interval is significant relative to sample thickness and may arise from zeroing drift, variation in mirror placement after re-seating, or calibration limitations. Zeroing drift and calibration accuracy can be improved through design refinement, whereas mirror placement variability may originate from imperfections in the shim samples. Although thickness variation was accounted for through the micrometer measurements, burrs or raised edges produced during manufacturing may alter the settled mirror position and reduce repeatability, particularly for nominally 0.5 mm samples. Future work should quantify the influence of these surface features or use {shims} with improved edge quality.

Table~\ref{tab:Results} also shows substantially larger errors and confidence intervals for nominally 0.5 mm samples. In the 0.3 mm calibration case, average error exceeded the 95\% confidence interval, indicating systematic bias. This behavior results from beam displacement exceeding the sensor’s linear response range. For this reason, calibration using a 0.5 mm shim was not attempted, as resulting measurements would not reflect true system capability. This limitation originates from the selected beam reflection angle. Future designs may reduce this angle to limit beam displacement for thicker samples, at the cost of increased sensitivity to calibration and zeroing errors.

\subsection{Silicone Fabrication Demonstration}

Silicone fabrication results, shown in Table~\ref{tab:SiliconeResults}, demonstrate that the spin-coating system can produce silicone sheets within 9~\unit{\micro\meter} of target thickness, based on micrometer measurements, satisfying the system design objectives.

The measurement system exhibited overall larger uncertainty than in the shim experiment, with a worst-case 95\% confidence interval of 58~\unit{\micro\meter}. A likely contributor is the Ecoflex10T material remaining slightly tacky when cured at room temperature, even after 24 hours. Unlike metal shims, the mirror cannot freely settle while in contact with the sample, increasing dependence on the mirror re-seating process between measurements and reducing repeatability. This behavior indicates that further refinement is required to improve measurement consistency.

Confidence intervals for the micrometer measurements were on the order of 10-25~\unit{\micro\meter}. This variability may result from deformation under measurement load. Despite the use of a plate stackup to distribute loading, the axial loading model does not perfectly represent actual measurement conditions: localized deformation between micrometer anvils will be greater than predicted, and applied measurement force is unlikely to remain constant between trials. Both effects contribute to micrometer measurement uncertainty.

\begin{table*}[htbp]
    \caption{Average measurements for silicone samples manufactured and measured via the spin coating system, with 95\% confidence intervals. Micrometer measurements have a $19.5$~\unit{\micro\meter} compensation factor added to account for deformation under measurement load, as derived from~(\ref{eqn:axialLoading}). For these silicone samples, the same calibration as the 0.3 mm shim samples was used.}%
    \begin{center}
    \begin{tabular}{|c|c|c|}
    \hline
    \textbf{Target (mm)} & \textbf{System Measurement (mm)} & \textbf{Micrometer Measurement (mm)} \\
    % \cline{2-4} 
    \hline
    0.3 & 0.299 ± 0.058 & 0.308 ± 0.024 \\
    0.3 & 0.296 ± 0.039 & 0.309 ± 0.012 \\ 
    0.3 & 0.302 ± 0.014 & 0.306 ± 0.021 \\ 
    \hline
    \end{tabular}
    \label{tab:SiliconeResults}
    \end{center}
\end{table*}

Additional error may arise from misalignment between the sample plate normal and the spin coater rotation axis, which can introduce thickness non-uniformity and increase micrometer measurement variance. Because the optical system measures displacement through a mirror resting on the sample surface, it cannot directly resolve local surface irregularities.

\section{Conclusion}
This work presents a low-cost, primarily 3D-printed spin coating platform with an integrated optical thickness measurement system for fabrication and characterization of highly compliant thin films. The system is designed to produce films in the 50–300~\unit{\micro\meter} range with repeatability within 10~\unit{\micro\meter}, addressing applications such as dielectric elastomer actuators where performance scales strongly with thickness. By combining fabrication and minimally deforming optical measurement in a single benchtop device, the platform enables in-situ thickness verification without sample removal or significant deflection, reducing handling- and contact-induced error.

Experimental validation demonstrated a thickness resolution of 3.6–3.7~\unit{\micro\meter} and best-case measurement repeatability of 13~\unit{\micro\meter} (95\% confidence interval), with silicone films manufactured within 9~\unit{\micro\meter} of target thickness. These results establish a quantitatively validated proof of concept and provide a foundation for iterative refinement toward more robust \nobreak{fabrication–measurement} systems that reduce reliance on specialized industrial metrology equipment.

\section*{Acknowledgments}
The authors thank The Japan Steel Works, LTD for their technical and financial support.

\bibliographystyle{asmems4}
\bibliography{references}

@article{de_groot_principles_2015,
	title = {Principles of interference microscopy for the measurement of surface topography},
	volume = {7},
	copyright = {https://doi.org/10.1364/OA\_License\_v1\#VOR},
	issn = {1943-8206},
	url = {https://opg.optica.org/abstract.cfm?URI=aop-7-1-1},
	doi = {10.1364/AOP.7.000001},
	language = {en},
	number = {1},
	urldate = {2026-02-16},
	journal = {Advances in Optics and Photonics},
	author = {De Groot, Peter},
	month = mar,
	year = {2015},
	pages = {1},
}

@book{fujiwara_spectroscopic_2007,
	edition = {1},
	title = {Spectroscopic {Ellipsometry}: {Principles} and {Applications}},
	copyright = {http://doi.wiley.com/10.1002/tdm\_license\_1.1},
	isbn = {978-0-470-01608-4 978-0-470-06019-3},
	shorttitle = {Spectroscopic {Ellipsometry}},
	url = {https://onlinelibrary.wiley.com/doi/book/10.1002/9780470060193},
	doi = {10.1002/9780470060193},
	language = {en},
	urldate = {2026-02-16},
	publisher = {Wiley},
	author = {Fujiwara, Hiroyuki},
	month = jan,
	year = {2007},
}

@article{heavens_optical_1960,
	title = {Optical properties of thin films},
	volume = {23},
	issn = {00344885},
	url = {https://iopscience.iop.org/article/10.1088/0034-4885/23/1/301},
	doi = {10.1088/0034-4885/23/1/301},
	number = {1},
	urldate = {2026-02-16},
	journal = {Reports on Progress in Physics},
	author = {Heavens, O S},
	month = jan,
	year = {1960},
	pages = {1--65},
}

@article{lawrence_mechanics_1988,
	title = {The mechanics of spin coating of polymer films},
	volume = {31},
	issn = {0031-9171},
	url = {https://pubs.aip.org/pfl/article/31/10/2786/951800/The-mechanics-of-spin-coating-of-polymer-films},
	doi = {10.1063/1.866986},
	language = {en},
	number = {10},
	urldate = {2026-02-16},
	journal = {The Physics of Fluids},
	author = {Lawrence, C. J.},
	month = oct,
	year = {1988},
	pages = {2786--2795},
}

@article{flack_mathematical_1984,
	title = {A mathematical model for spin coating of polymer resists},
	volume = {56},
	issn = {0021-8979, 1089-7550},
	url = {https://pubs.aip.org/jap/article/56/4/1199/171380/A-mathematical-model-for-spin-coating-of-polymer},
	doi = {10.1063/1.334049},
	language = {en},
	number = {4},
	urldate = {2026-02-16},
	journal = {Journal of Applied Physics},
	author = {Flack, Warren W. and Soong, David S. and Bell, Alexis T. and Hess, Dennis W.},
	month = aug,
	year = {1984},
	pages = {1199--1206},
}

@article{meyerhofer_characteristics_1978,
	title = {Characteristics of resist films produced by spinning},
	volume = {49},
	issn = {0021-8979, 1089-7550},
	url = {https://pubs.aip.org/jap/article/49/7/3993/505461/Characteristics-of-resist-films-produced-by},
	doi = {10.1063/1.325357},
	language = {en},
	number = {7},
	urldate = {2026-02-16},
	journal = {Journal of Applied Physics},
	author = {Meyerhofer, Dietrich},
	month = jul,
	year = {1978},
	pages = {3993--3997},
}

@misc{chemat_chemat_nodate,
	title = {Chemat precision spin-coater},
	url = {https://www.sigmaaldrich.com/US/en/product/aldrich/z551589},
	language = {en},
	urldate = {2026-02-13},
	journal = {Sigma Aldrich},
	author = {{Chemat}},
}

@misc{alex_important_2024,
	title = {Important {Update}: {BLHeli}\_32 is {Shutting} {Down}},
	shorttitle = {{BLHeli}\_32 is {Shutting} {Down}},
	url = {https://www.dronetrest.com/t/important-update-blheli-32-is-shutting-down/10509},
	language = {en},
	urldate = {2026-02-13},
	journal = {DroneTrest},
	author = {{Alex}},
	month = jun,
	year = {2024},
}

@misc{bruker_filmtek_nodate,
	title = {{FilmTek} {SE}},
	url = {https://www.bruker.com/en/products-and-solutions/test-and-measurement/ellipsometers-and-reflectometers/spectroscopic-ellipsometers/filmtek-se.html},
	language = {en},
	urldate = {2026-02-13},
	journal = {Bruker},
	author = {{Bruker}},
}

@misc{ossila_spin_nodate,
	title = {Spin {Coater}},
	url = {https://www.ossila.com/products/spin-coater},
	language = {en},
	urldate = {2026-02-13},
	journal = {Ossila},
	author = {{Ossila}},
}

@misc{preasion_spin_nodate,
	title = {Spin {Coater} {Laboratory} {Super} {Spin} {Coater} {Vacuum} {Glue} {Coating} {Homogenizer} {Spin} {Coating} {Instrument} 100-{8000RPM}},
	url = {https://www.walmart.com/ip/Spin-Coater-Laboratory-Super-Spin-Coater-Vacuum-Glue-Coating-Homogenizer-Spin-Coating-Instrument-100-8000RPM/5450790017},
	language = {en-US},
	urldate = {2026-02-13},
	journal = {Walmart.com},
	author = {{Preasion}},
}

@techreport{noauthor_pla_nodate,
	type = {Technical {Data} {Sheet}},
	title = {{PLA} {Basic}},
	url = {https://store.bblcdn.com/s1/default/58b85d0f3db94878854a28fdb8a0006e/Bambu_PLA_Basic_Technical_Data_Sheet.pdf},
	number = {3.0},
	urldate = {2026-02-02},
	institution = {Bambu Lab},
}

@article{emslie_flow_1958,
	title = {Flow of a {Viscous} {Liquid} on a {Rotating} {Disk}},
	volume = {29},
	issn = {0021-8979, 1089-7550},
	url = {https://pubs.aip.org/jap/article/29/5/858/162136/Flow-of-a-Viscous-Liquid-on-a-Rotating-Disk},
	doi = {10.1063/1.1723300},
	language = {en},
	number = {5},
	urldate = {2026-01-29},
	journal = {Journal of Applied Physics},
	author = {Emslie, Alfred G. and Bonner, Francis T. and Peck, Leslie G.},
	month = may,
	year = {1958},
	pages = {858--862},
}

@article{ferreira_experimental_2017,
	title = {Experimental characterization and micrography of {3D} printed {PLA} and {PLA} reinforced with short carbon fibers},
	volume = {124},
	issn = {13598368},
	url = {https://linkinghub.elsevier.com/retrieve/pii/S135983681633195X},
	doi = {10.1016/j.compositesb.2017.05.013},
	language = {en},
	urldate = {2026-01-29},
	journal = {Composites Part B: Engineering},
	author = {Ferreira, Rafael Thiago Luiz and Amatte, Igor Cardoso and Dutra, Thiago Assis and Bürger, Daniel},
	month = sep,
	year = {2017},
	pages = {88--100},
}

@article{carbonell_rubio_maasi_2022,
	title = {Maasi: {A} {3D} printed spin coater with touchscreen},
	volume = {11},
	issn = {24680672},
	shorttitle = {Maasi},
	url = {https://linkinghub.elsevier.com/retrieve/pii/S246806722200061X},
	doi = {10.1016/j.ohx.2022.e00316},
	language = {en},
	urldate = {2025-07-28},
	journal = {HardwareX},
	author = {Carbonell Rubio, Dani and Weber, Willi and Klotzsch, Enrico},
	month = apr,
	year = {2022},
	pages = {e00316},
}

@inproceedings{jamali_development_2022-1,
	address = {Hilton Head, South Carolina, USA},
	title = {{DEVELOPMENT} {OF} {A} {SCALABLE} {SOFT} {FINGER} {GRIPPER} {FOR} {SOFT} {ROBOTS}},
	isbn = {978-1-940470-04-7},
	url = {https://transducer-research-foundation.org/technical_digests/HiltonHead_2022/hh2022_0352.pdf},
	doi = {10.31438/trf.hh2022.88},
	language = {en},
	urldate = {2025-06-18},
	booktitle = {2022 {Solid}-{State}, {Actuators}, and {Microsystems} {Workshop} {Technical} {Digest}},
	publisher = {Transducer Research Foundation},
	author = {Jamali, Armin and Knoerlein, Robert and Goldschmidtboeing, Frank and Woias, Peter},
	month = jun,
	year = {2022},
	pages = {352--355},
}

@inproceedings{muffoletto_anticipating_2013,
	address = {San Diego, California, USA},
	title = {Anticipating electrical breakdown in dielectric elastomer actuators},
	volume = {8687},
	issn = {0277-786X},
	url = {http://proceedings.spiedigitallibrary.org/proceeding.aspx?doi=10.1117/12.2009932},
	doi = {10.1117/12.2009932},
	language = {en},
	urldate = {2025-07-15},
	booktitle = {{SPIE} {Proceedings}},
	publisher = {SPIE},
	author = {Muffoletto, Daniel P. and Burke, Kevin M. and Zirnheld, Jennifer L.},
	editor = {Bar-Cohen, Yoseph},
	month = apr,
	year = {2013},
	pages = {86870U},
}

\appendices

\section{Full off-the-shelf component list for spin coater and measurement System}
\label{appendix:coaterParts}

Please note that the following component lists in Tables~\ref{tab:SpinCoaterComponents} and \ref{tab:MeasurementComponents} do not include hardware such as screws, nuts, washers, or common electrical components such as wires or resistors.
\begin{table}[htbp]
    \caption{Spin coater hardware and control components.}
    \centering
    \begin{tabular}{|c|p{2.9cm}|p{3.4cm}|}
    \hline
    \textbf{Qty.} & \textbf{Component} & \textbf{Source / Part ID} \\
    \hline
    1 & SPARK MAX Motor Controller & REV Robotics, REV-11-2158 \\
    \hline
    1 & NEO 550 Brushless Motor & REV Robotics, REV-21-1651 \\
    \hline
    1 & UltraPlanetary 550 Pinion Gear (13T) & REV Robotics, REV-41-1608-PK2 \\
    \hline
    1 & Arduino UNO R4 Minima & Arduino, ABX00080 \\
    \hline
    1 & MCP2515 SPI CAN Bus Module & Amazon, B01D0WSEWU \\
    \hline
    6+ & Neodymium Magnets (1/4" OD $\times$ 1/8" thick) & McMaster-Carr, 58605K33 \\
    \hline
    7 & M3 Heat-Set Threaded Inserts & -- \\
    \hline
    \end{tabular}
    \label{tab:SpinCoaterComponents}
\end{table}
\begin{table}[htbp]
    \caption{Optical measurement system components.}
    \centering
    \begin{tabular}{|c|p{2.9cm}|p{3.4cm}|}
    \hline
    \textbf{Qty.} & \textbf{Component} & \textbf{Source / Part ID} \\
    \hline
    1 & 895 nm, 0.2 mW Laser Diode & ThorLabs, L895VH1 \\
    \hline
    1 & 0.5" Diam. Lens Tube & ThorLabs, SM05L03 \\
    \hline
    1 & 0.5" Diam. Plano-Convex Lens, 20 mm FL & ThorLabs, LA1074-AB \\
    \hline
    1 & XY Linear Stage & Newport, DS40-XY \\
    \hline
    1 & Quadrant Detector Sensor & ThorLabs, PDQ80A \\
    \hline
    1 & SM05--SM1 Adapter & ThorLabs, SM1A1 \\
    \hline
    1 & SM1 Lens Tube & ThorLabs, SM1L03 \\
    \hline
    1 & 1" Diam. Plano-Convex Lens, 25.4 mm FL & ThorLabs, LA1951-AB \\
    \hline
    1 & myRIO-1900 & National Instruments \\
    \hline
    1 & Mirror Material & Amazon, B01KU7XVVC \\
    \hline
    1 & 0.2 mm Shims & McMaster, 98089A630 \\
    \hline
    1 & 0.3 mm Shims & McMaster, 98089A742 \\
    \hline
    1 & 0.5 mm Shims & McMaster, 98089A854 \\
    \hline
    \end{tabular}
    \label{tab:MeasurementComponents}
\end{table}

\section{Print Settings for Custom PLA Components} \label{appendix:printSettings}
All 3D printed components were printed with a Bambu Lab X1 Carbon, and thus all modified print settings will be listed using the terminology that is used in Bambu Studio, Bambu Lab's slicer software. The base profile used was ``.20mm Standard @BBL X1C" with the default material preset ``PLA Basic." All settings not mentioned in the following list were unchanged from these presets.

All parts were printed with Bambu Lab PLA Basic, but Bambu Support for PLA was used via AMS filament system for supports. Note that, at time of writing, Bambu Studio will generate a pop-up notification when selecting Support for PLA as a material for the support interface. This pop-up suggests print setting changes that are not used, and thus should not be accepted.
\begin{itemize}
    \item Wall Loops: 5
    \item Top Shell Layers: 5
    \item Bottom Shell Layers: 5
    \item Sparse Infill Density: 25\%
    \item Sparse Infill Pattern: Gyroid
    \item Enable Support: True
    \item Support Type: tree(auto)
    \item Support Style: Default
    \item Threshold Angle: 30\degree
    \item On build plate only: True
    \item Remove Small Overhangs: False
    \item Filament for Supports Support/Raft Interface: Bambu Support for PLA
\end{itemize}

\section{Derivation of Beam Motion Scaling based on Angle of Reflection} \label{appendix:beamProof}

Note that this appendix refers to variables from Fig.~\ref{fig:beamGeometry}.

\begin{align}
     \tag{A1} \phi & = \frac{\pi}{2} - \theta\\
     \tag{A2} \tau & = m \sin(\phi) \\
     \tag{A3} m & = \frac{\tau}{\sin(\phi)} \\
     \tag{A4} y & = m \sin(2\phi) \label{eqn:doubleAngle}\\
     \tag{A5} y & = \frac{\sin(2\phi)}{\sin(\phi)}\tau \\
     \tag{\ref{eqn:beamDeflection}} y & = \frac{\sin(\pi - 2\theta)}{\sin(\frac{\pi}{2} - \theta)} \tau
\end{align}
\end{document}